\documentclass[10pt,conference,a4paper]{IEEEtran}

\usepackage{cite}
\usepackage[pdftex]{graphicx}
\usepackage{amsmath}
\usepackage{bm}
\usepackage{color}
\usepackage[table]{xcolor}

\begin{document}

\title{q-SNE: Visualizing Data using q-Gaussian Distributed Stochastic Neighbor Embedding}

\author{\IEEEauthorblockN{Motoshi Abe}
\IEEEauthorblockA{\textit{School of Engineering} \\
\textit{Hiroshima University}\\
Higashi-hiroshima, Japan \\
i13abemotoshi@gmail.com}
\and
\IEEEauthorblockN{Junichi Miyao}
\IEEEauthorblockA{\textit{Department of Information Engineering} \\
\textit{Hiroshima University}\\
Higashi-hiroshima, Japan \\
miyao@hiroshima-u.ac.jp}
\and
\IEEEauthorblockN{Takio Kurita}
\IEEEauthorblockA{\textit{Department of Information Engineering} \\
\textit{Hiroshima University}\\
Higashi-hiroshima, Japan \\
tkurita@hiroshima-u.ac.jp}}

\maketitle

\begin{abstract}
The dimensionality reduction has been widely introduced to use the high-dimensional data for regression, classification, feature analysis, and visualization.
As the one technique of dimensionality reduction, a stochastic neighbor embedding (SNE) was introduced.
The SNE leads powerful results to visualize high-dimensional data by considering the similarity between the local Gaussian distributions of high and low-dimensional space.
To improve the SNE, a t-distributed stochastic neighbor embedding (t-SNE) was also introduced.
To visualize high-dimensional data, the t-SNE leads to more powerful and flexible visualization on 2 or 3-dimensional mapping than the SNE by using a t-distribution as the distribution of low-dimensional data.
Recently, Uniform manifold approximation and projection (UMAP) is proposed as a dimensionality reduction technique.
We present a novel technique called a q-Gaussian distributed stochastic neighbor embedding (q-SNE).
The q-SNE leads to more powerful and flexible visualization on 2 or 3-dimensional mapping than the t-SNE and the SNE by using a q-Gaussian distribution as the distribution of low-dimensional data.
The q-Gaussian distribution includes the Gaussian distribution and the t-distribution as the special cases with q=1.0 and q=2.0.
Therefore, the q-SNE can also express the t-SNE and the SNE by changing the parameter q, and this makes it possible to find the best visualization by choosing the parameter q.
We show the performance of q-SNE as visualization on 2-dimensional mapping and classification by k-Nearest Neighbors (k-NN) classifier in embedded space compared with SNE, t-SNE, and UMAP by using the datasets MNIST, COIL-20, OlivettiFaces, FashionMNIST, and Glove.
\end{abstract}

\IEEEpeerreviewmaketitle

\section{Introduction}
With the development of IT technology, the amount of data to be processed is increasing rapidly, and it is becoming common to process high-dimensional data accordingly.
Since we can only recognize samples in 2 or 3-dimensional space, it is difficult to understand the distribution of the samples in the high-dimensional space. 
To see the samples in the high-dimensional space, we have to utilize unsupervised or supervised dimensionality reduction methods that get the low-dimensional approximation of the samples in the high-dimensional space.
Also, the dimensionality reduction has been widely used in regression or classification for pre-processing.

Many techniques are available to reduce the dimensions of the data.
Examples of the unsupervised dimensionality reduction techniques are (PCA)\cite{pearson1901liii, kurita2014principal}, Kernel PCA (KPCA)\cite{scholkopf1997kernel}, Stochastic neighbor embedding (SNE)\cite{hinton2003stochastic}, t-distributed SNE (t-SNE)\cite{maaten2008visualizing}, Sammon mapping\cite{sammon1969nonlinear}, Isomap\cite{tenenbaum2000global}, Linear Embedding (LLE)\cite{roweis2000nonlinear}, Maximum Variance Unfolding (MVU)\cite{weinberger2004learning}, Laplacian Eigenmaps\cite{belkin2002laplacian}, curvilinear components analysis (CCA)\cite{demartines1997curvilinear}, and Uniform manifold approximation and projection (UMAP)\cite{mcinnes2018umap}.
Other dimensionality reduction techniques are reviewed in \cite{liu2016visualizing}

The visualization of high-dimensional data is very important to understand the relationship between the samples in a given dataset.
For visualization of the high-dimensional data, it is also necessary to reduce the dimension of the data and we can use the techniques for these dimensionality reduction techniques.
Recently, the SNE and t-SNE made a great contribution to the visualization of the high-dimensional data by embedding the distributions in the high-dimensional data space into the low-dimensional space.

A stochastic neighbor embedding (SNE)\cite{hinton2003stochastic} has been introduced as a technique for dimensionality reduction.
The SNE embeds the similarity between samples in the high-dimensional space which are defined by using local Gaussian distribution into the low-dimensional space which are also defined by using local Gaussian distribution.
The Kullback-Leibler divergence between the local Gaussian distributions of the original high-dimensional space and the embedded low-dimensional space is used to measure the goodness of the embedded space.
It is known that the SNE can visualize the distributions of the high-dimensional data in 2 or 3-dimensional embedded space.
However, the SNE has some problems in which the Gaussian distribution in the embedded low-dimensional space does not give enough weights for the distant samples from the center point and the separation between the clusters or the samples is not enough in the embedded space.

To address these problems, t-distributed stochastic neighbor embedding (t-SNE)\cite{maaten2008visualizing} has been proposed as an extension of the SNE for dimensionality reduction.
The t-SNE uses the local Gaussian distribution in the high-dimensional space but the local t-distribution is used in low-dimensional space instead of the local Gaussian distribution in the SNE.
Since the kurtosis of the t-distribution is larger than the Gaussian distribution, it is expected that the separation between the clusters of the samples in the embedded low-dimensional space can be improved than the SNE.
It is known that t-SNE can propose a more understandable plot of the samples in the low-dimensional embedded space in which near samples become close and distant samples become far.
Also, the t-SNE has been improved for computation by using a tree-based algorithm\cite{van2014accelerating}.
Thus, the t-SNE is often used as one of the standard tools for visualization \cite{platzer2013visualization, rauber2016visualizing, li2017application}.
Not only visualization tool, but also t-SNE is used in \cite{xu2019d}.

Recently, Uniform manifold approximation and projection (UMAP)\cite{mcinnes2018umap} is proposed as a dimensionality reduction technique.
The UMAP intends to model the manifold with a fuzzy topological structure and embeds the fuzzy topological structure of the high-dimensional space to the low-dimensional space.

In this paper, we propose an extension of the t-SNE by using q-Gaussian distribution\cite{tanaka2019entropy} instead of t-distribution in the t-SNE.
We call the proposed method q-Gaussian distributed stochastic neighbor embedding (q-SNE).
The q-Gaussian distribution is derived by the maximization of the Tsallis entropy under appropriate constraints and is a generalization of the Gaussian distribution.
The q-Gaussian distribution has the parameter $q$ and we can recover the Gaussian distribution by setting the parameter $q \to 1$ in the q-Gaussian distribution.
Also, the t-distribution can be recovered by setting the parameter $q=2.0$.
Similar to the original SNE or the t-SNE, we use the local Gaussian distribution in high-dimensional space.
On the other hand, the local q-Gaussian distribution in low-dimensional space is used in the proposed q-SNE instead of the local Gaussian distribution in the SNE or the local t-distribution in the t-SNE.
Since the q-Gaussian distribution is an extension of the Gaussian distribution and the t-distribution with parameter $q$, the proposed q-SNE can generate the same low-dimensional embedded space with the SNE or the t-SNE by changing the parameter $q$.
Also, we can generate the embedded space with a more separated plot than the t-SNE by taking the parameter $q > 2.0$.
We think that this flexibility to construct the embedded space can enable us to make the user-friendly visualization tool.
Not only visualization tools but also show the effectiveness of classification accuracy by choosing the parameter $q$.

We show the flexibility of the q-SNE as visualization on 2-dimensional mapping and classification by k-Nearest Neighbors (k-NN) classifier in embedded space compared with SNE, t-SNE and UMAP by using MNIST,  COIL-20, OlivettiFaces, FashionMNIST, and GloVe as dataset.

\section{Related Works}
\subsection{SNE}
The SNE embeds the pairwise similarities between samples in the high-dimensional space into the low-dimensional space.
The goodness of the low-dimensional space is evaluated as the Kullback-Leibler divergence between the conditional probabilities of the samples in the high-dimensional space and the low-dimensional space.
Both the conditional probabilities in the high-dimensional space and the low-dimensional space are defined by using local Gaussian distribution.

Let $\{\bm{x}_i|i=1\ldots N\}$ be a set of the samples in the high-dimensional space.
We assume that the vectors of each samples are represented as $\bm{x}_i = \begin{bmatrix} x_{i1} & x_{i2} & \cdots & x_{iD} \end{bmatrix}^T$ and the dimension of the vector is $D>2$.

To define the pairwise similarities between samples in the high and low-dimensional space, we define the conditional probability by using the local Gaussian distribution.
The conditional probability in the high-dimensional space is defined as
\begin{align}
    \label{pij}
    p_{j|i} = \frac{\exp{(-\|\bm{x}_i-\bm{x}_j\|^2/2\sigma_i^2)}}{\sum_{k\neq i}^N \exp{(-\|\bm{x}_i-\bm{x}_k\|^2/2\sigma_i^2)}},
\end{align}
where $\sigma_i$ is the variance of the local Gaussian distribution around sample $\bm{x}_i$ which is determined by binary search by using the entropy defined as
\begin{align}
    \label{perp}
    \log{k} = -\sum_{j \neq i}^N p_{j|i}\log{p_{j|i}},
\end{align}
where $k$ is called perplexity.
Eq.(\ref{pij}) defines the local Gaussian distribution in high-dimensional space for all samples around the sample $\bm{x}_i$, and $p_{i|i}$ is set to be $0$ because we are interested in only the pairwise similarities.

Let $\{\bm{y}_i|i=1\ldots N\}$ be a set of the embedded vectors in the low-dimensional space of the samples $\{\bm{x}_i|i=1,\ldots N\}$.
The vectors in the embedded low-dimensional space are represented as $\bm{y}_i = \begin{bmatrix} y_{i1} & \cdots & y_{id} \end{bmatrix}^T$ and the dimension of the low-dimensional space is much smaller than the original space as $d<D$.

Similarly, the conditional probability in the embedded low-dimensional space is defined as
\begin{align}
    \label{rij}
    r_{j|i} = \frac{\exp{(-\|\bm{y}_i-\bm{y}_j\|^2)}}{\sum_{k\neq i}^N \exp{(-\|\bm{y}_i-\bm{y}_k\|^2)}},
\end{align}
where $r_{i|i}$ is also set to be $0$.
Eq.(\ref{rij}) defines the local Gaussian distribution in the embedded low-dimensional space for all samples around the sample $\bm{y}_i$.

The Kullback-Leibler divergence between these conditional probabilities in the original high-dimensional space and the embedded low-dimensional space is used to measure the goodness of the embedded space and is maximized to obtain the vectors in the embedded space.
The Kullback-Leibler divergence is defined as
\begin{align}
    \label{KL}
    C = \sum_i^N\sum_{j\neq i}^N p_{j|i}\log{\frac{p_{j|i}}{r_{j|i}}}.
\end{align}

The SNE finds the embedded vectors $\{\bm{y}_i\}$ in the low-dimensional space of the samples $\{\bm{x}_i\}$ in the high-dimensional space by minimizing the Kullback-Leibler divergence $C$.
The update rule of $\bm{y}_i$ by the gradient decent is given as
\begin{align}
    \label{optim}
    \bm{y}^{t+1}_i=\bm{y}^t_i-\eta \frac{\partial C}{\partial \bm{y}_i}+\alpha (t)(\bm{y}^t_i-\bm{y}^{t-1}_i),
\end{align}
where $t$, $\eta$, $\alpha (t)$, and $\frac{\partial C}{\partial \bm{y}_i}$ are respectively the iteration,  the learning rate, the momentum of iteration $t$, and the gradient defined as
\begin{align}
    \label{SNEgrad}
    \frac{\partial C}{\partial \bm{y}_i} = 2\sum_j^N(p_{j|i}-r_{j|i}+p_{i|j}-r_{i|j})(\bm{y}_i-\bm{y}_j).
\end{align}

Hinton et al. proposed the symmetric SNE in \cite{maaten2008visualizing}.
The symmetric SNE uses joint probability instead of the conditional probability in the original SNE.
The joint probability in the high-dimensional space is defined as
\begin{align}
    \label{sympij}
    p_{ij} = \frac{1}{2}(p_{i}p_{j|i}+p_{j}p_{i|j}) = \frac{p_{j|i}+p_{i|j}}{2N},
\end{align}
where $p_{i}=p_{j}=\frac{1}{N}$, $p_{ii}$ is $0$, and $p_{ij}=p_{ji}$ for $\forall i,j$.
Similarly, the joint probability in the low-dimensional space is define as
\begin{align}
    \label{symrij}
    r_{ij} = \frac{\exp{(-\|\bm{y}_i-\bm{y}_j\|^2)}}{\sum_l^N\sum_{k\neq l}^N \exp{(-\|\bm{y}_l-\bm{y}_k\|^2)}},
\end{align}
where $r_{ii}$ is $0$, and $r_{ij}=r_{ji}$ for $\forall i,j$.

Then, the Kullback-Leibler divergence is defined as
\begin{align}
    \label{symKL}
    C = \sum_i^N\sum_{j\neq i}^N p_{ij}\log{\frac{p_{ij}}{r_{ij}}}.
\end{align}
The optimization is performed by using the same equation with Eq.(\ref{optim}) and the gradient for this case becomes more simple and is defined as
\begin{align}
    \label{symSNEgrad}
    \frac{\partial C}{\partial \bm{y}_i} = 4\sum_j^N(p_{ij}-r_{ij})(\bm{y}_i-\bm{y}_j).
\end{align}

\subsection{t-SNE}
In the SNE or the symmetric SNE, the conditional probability or joint probability in the low-dimensional space is defined by using the local Gaussian distribution.
However, the separation between the clusters or the samples is not enough in the embedded space because the Gaussian distribution can not give enough weights for the distant samples from the center point.

To improve these problems, the t-SNE has been proposed as an extension of the SNE.
The t-SNE uses the local t-distribution in low-dimensional space instead of the local Gaussian distribution in the SNE.
Since the kurtosis of the t-distribution is larger than the Gaussian distribution, it is expected that the separation between the clusters of the samples in the embedded low-dimensional space by the t-SNE can be improved than the SNE.

The joint probability in the high-dimensional space is defined by using the local Gaussian distribution similar to the symmetric SNE.
The joint probability in the embedded low-dimensional space is defined by using t-distribution as
\begin{align}
    \label{trij}
    r_{ij}=\frac{(1+\|\bm{y}_i-\bm{y}_j\|^2)^{-1}}{\sum_l^N\sum_{k\neq l}^N(1+\|\bm{y}_k-\bm{y}_l\|^2)^{-1}},
\end{align}
where $r_{ii}$ is $0$, and $r_{ij}=r_{ji}$ for $\forall i,j$.
The Kullback-Leibler divergence and the update rule for optimization are almost the same as Eq.(\ref{symKL}) and Eq.(\ref{optim}).
The gradient for the t-SNE is given as
\begin{align}
    \label{tSNEgrad}
    \frac{\partial C}{\partial \bm{y}_i} = 4\sum_j^N(p_{ij}-r_{ij})(\bm{y}_i-\bm{y}_j)(1+\|\bm{y}_i-\bm{y}_j\|^2)^{-1}.
\end{align}

It is known that the t-SNE can produce a more understandable plot of the samples in the low-dimensional embedded space.

\subsection{UMAP}
Recently, the UMAP has been proposed as a dimensionality reduction technique.
They mention hyperparameters to control embedding as follows,
\begin{itemize}
    \item $nn$, the number of neighbors to consider when approximating the local metric;
    \item $d$, the target embedding dimension;
    \item $min\_dist$, the desired separation between close points in the embedding
space; and
    \item $n-epochs$, the number of training epochs to use when optimizing the low
dimensional representation.
\end{itemize}

To compare with our proposed method, we used the UMAP for experiments.

\section{q-Gasussian Distributed Stochastic Neighbor Embedding}

\begin{figure}[t!]
    \centering
    \includegraphics[width=0.8\linewidth]{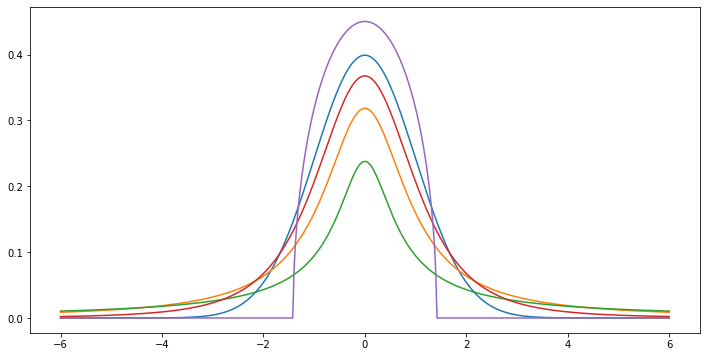}\\
    \includegraphics[width=\linewidth]{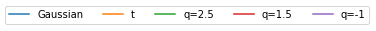}
    \caption{This figure shows the graphs about a Gaussian distribution, a t-distribution, and some q-Gaussian distributions. Gaussian denotes the Gaussian distribution as blue line. t denotes the t-distribution of degrees of freedom $1$ as orange line. q=2.5, q=1.5, and q=-1 denote the q-Gaussian distributions as green, red, and purple line, respectively.}
    \label{graph}
\end{figure}

We propose an extension of the t-SNE by using q-Gaussian distribution\cite{tanaka2019entropy} instead of t-distribution in the t-SNE.
We call the proposed method q-Gaussian distributed stochastic neighbor embedding (q-SNE).

The motivation for proposing the q-SNE is that the q-Gaussian distribution can express as Gaussian distribution, t-distribution, and others depending on the value of q, so we can expect a distribution that allows for better visualization.

\subsection{q-Gaussian Distribution}
The q-Gaussian distribution is derived by the maximization of the Tsallis entropy under appropriate constraints and is a generalization of the Gaussian distribution.

Let $s$ be a 1-dimensional observation.
The q-Gaussian distribution for the observation $s$ is defined as
\begin{align}
    \label{qdist}
    P_q(s;\mu,\sigma^2)=\frac{1}{Z_q}\left(1+\frac{q-1}{3-q}\frac{(s-\mu)^2}{\sigma^2}\right)^{-\frac{1}{q-1}}
\end{align}
where $\mu$ and $\sigma$ are the mean and the variance, respectively.
The normalization factor $Z_q$ is given by 
\begin{align}
    \label{qcoef}
    Z_q = 
    \begin{cases}
    \sqrt{\frac{3-q}{q-1}}Beta\left(\frac{3-q}{2(q-1)},\frac{1}{2} \right)\sigma ,\hspace{0.8cm} &1\leq q < 3\\
    \vspace{0.2mm}\\
    \sqrt{\frac{3-q}{1-q}}Beta\left(\frac{2-q}{1-q},\frac{1}{2}\right)\sigma ,  &q < 1
    \end{cases}
\end{align}
where $Beta()$ is the beta function.
It is known that the q-Gaussian distribution defined by Eq.(\ref{qdist}) always satisfies the inequality 
\begin{align}
    \label{costraint}
    1 + \frac{q-1}{3-q}\frac{(s-\mu)^2}{\sigma^2}\geq 0.
\end{align}

The q-Gaussian distribution has the parameter $q$ as shown in Eq.(\ref{qdist}) and we can recover the Gaussian distribution and the t-distribution by setting the parameter $q$ in the q-Gaussian distribution.
Fig.\ref{graph} shows the graph of the Gaussian distribution, the t-distribution, and the q-Gaussian distributions with a few different parameters $q$.
In this graph, we set $\mu$ and $\sigma$ to $0$ and $1$.
If $q \to 1$, then the q-Gaussian distribution becomes the Gaussian distribution.
If $q =1+\frac{2}{n+1}$, then the q-Gaussian distribution becomes the t-distribution of degrees of freedom $n$.
%
From Fig.\ref{graph}, it is noticed that the q-Gaussian distribution has a more sharp peak at $0$ than the t-distribution.

\subsection{q-SNE}
The q-SNE uses the q-Gaussian distribution in low-dimensional space instead of the local Gaussian distribution in the SNE or the local t-distribution in the t-SNE.
Similar to the symmetric SNE or the t-SNE, we use the local Gaussian distribution in high-dimensional space.
The joint probability in the low-dimensional space is defined as
\begin{align}
    \label{qrij}
    r_{ij}=\frac{(1+\frac{q-1}{3-q}\|\bm{y}_i-\bm{y}_j\|^2)^{-\frac{1}{q-1}})}{\sum_l^N\sum_{k\neq l}^N(1+\frac{q-1}{3-q}\|\bm{y}_l-\bm{y}_k\|^2)^{-\frac{1}{q-1}})},
\end{align}
where $q$ is a hyperparameter, $r_{ii}$ is $0$, and $r_{ij}=r_{ji}$ for $\forall i,j$.

The Kullback-Leibler divergence and the update rule for optimization are the same as Eq.(\ref{symKL}) and Eq.(\ref{optim}), respectively.
The gradient for $\bm{y}_i$ is given as
\begin{align}
    \label{qSNEgrad}
    &\frac{\partial C}{\partial \bm{y}_i} =\notag\\ &\frac{4}{3-q}\sum_j^N(p_{ij}-r_{ij})(\bm{y}_i-\bm{y}_j)(1+\frac{q-1}{3-q}\|\bm{y}_i-\bm{y}_j\|^2)^{-1}.
\end{align}

Since the q-Gaussian distribution is an extension of the Gaussian distribution and the t-distribution with the parameter $q$, the proposed q-SNE can generate the same low-dimensional embedded space with the SNE or the t-SNE by changing the parameter $q$.

\begin{figure}[t!]
    \centering
    \includegraphics[width=0.5\linewidth]{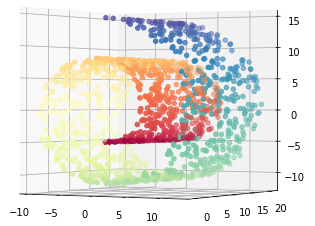}
    \caption{This figure shows the 1500 points of swissroll dataset in 3-dimensional mapping.}
    \label{swiss}
\end{figure}

\begin{figure}[tb]
    \centering
    \includegraphics[width=40mm]{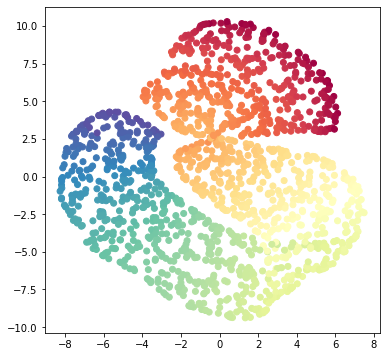}
    \includegraphics[width=40mm]{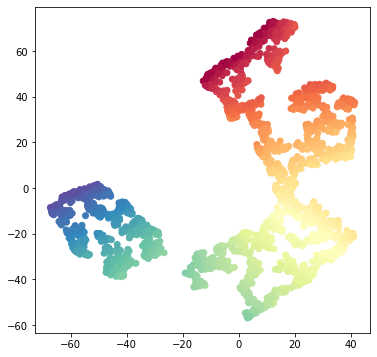}\\
    (a) SNE [G.Hinton]\hspace*{11mm}
    (b) t-SNE [L.Maaten]\\
    \includegraphics[width=40mm]{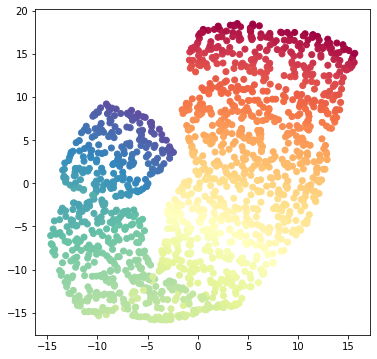}
    \includegraphics[width=40mm]{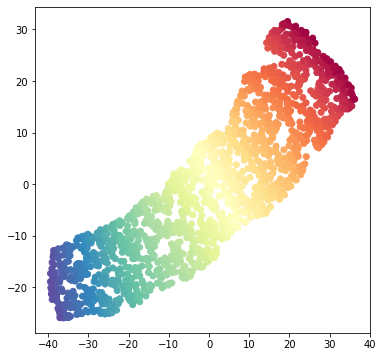}\\
    (c) $q=1.1$ \hspace*{25mm}
    (d) $q=1.5$\\
    \includegraphics[width=40mm]{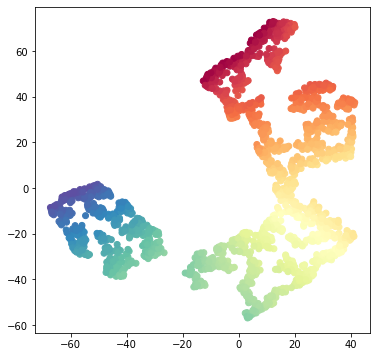}
    \includegraphics[width=40mm]{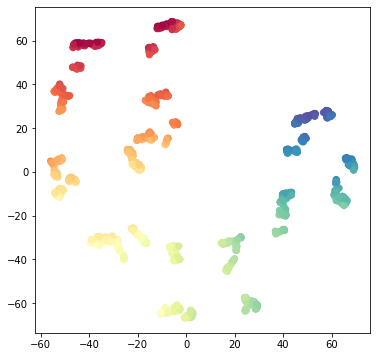}\\
    (e) $q=2.0$ \hspace*{25mm}
    (f) $q=2.5$\\
    \caption{Visualization of swissroll dataset in the 2-dimensional embedded space.}
    \label{qswiss}
\end{figure}

\section{Experiments}

\subsection{Preliminary Experiment using Swissroll Data}

To confirm the effectiveness of the proposed q-SNE, we have performed preliminary experiments using a 3-dimensional swissroll dataset shown in Fig.\ref{swiss} compared with SNE and t-SNE. 
The number of samples in this dataset is $1500$.

To calculate the joint probability $p_{ij}$ in the original 3-dimensional space, the variance $\sigma$ for each sample is determined by setting the perplexity to $30$.
In the optimization to obtain the embedded low-dimensional vectors of each sample, the update rule is applied $1000$ times starting from the random initial vectors. 
The learning rate is set to $200$ and 
the momentum is controlled such that it is set to $0.5$ for the first $250$ iterations and $0.8$ for the remaining iterations.
To speed up the optimization in the early stages, the joint probability in the high-dimensional space $p_{ij}$ is multiplied $12$ for the first $250$ iterations.
These settings are almost the same as the implementation of t-SNE in the scikit-learn \cite{pedregosa2011scikit}.

Fig.\ref{qswiss} shows the visualization of the samples in the embedded 2-dimensional space for SNE, t-SNE and q-SNE with $q=1.1$, $q=1.5$, $q=2.0$, and $q=2.5$.
Since q-Gaussian distribution becomes the same as Gaussian distribution when $q \to 1$, the visualization obtained by the q-SNE with $q=1.1$ is close to the result of the standard SNE.
Similarly, the visualization obtained by the q-SNE with $q=2.0$ is the same as the result of the t-SNE because q-Gaussian distribution becomes the same as the t-distribution.
From this figure, we can notice that a sheet of the swissroll is correctly visualized in Fig.\ref{qswiss} (d).
But the samples are more clustered in Fig.\ref{qswiss} (e) or (f).
These results show that the proposed q-SNE can visualize the samples in the 2-dimensional embedded space with different connectivity of the samples by changing the parameter $q$.
This makes it possible to find the best visualization by choosing the parameter $q$.

\begin{figure}[tb]
    \centering
    \includegraphics[width=42mm]{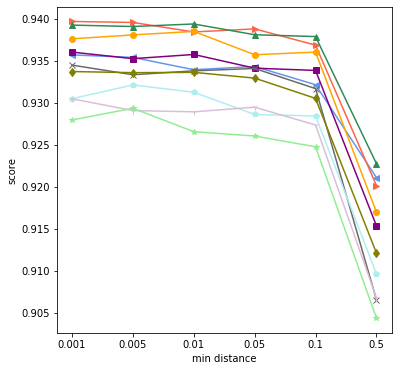}
    \includegraphics[width=42mm]{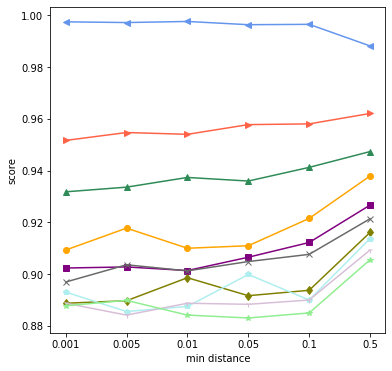}
    (a) MNIST \hspace*{25mm} (b) COIL-20\\
    \includegraphics[width=42mm]{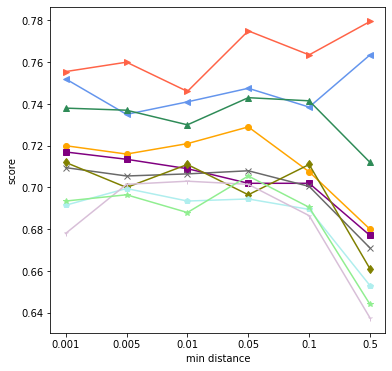}
    \includegraphics[width=42mm]{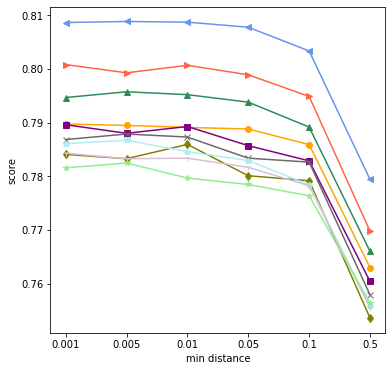} \\
    (c) OlivettiFaces \hspace*{15mm} (d) FAshionMNSIT\\
    \includegraphics[width=\linewidth]{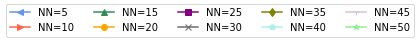}
    \caption{These graphs show the classification accuracy of UMAP by the k-NN classifier ($k=10$) in the embedding space. The x-axis denotes the $min\_dist$. The y-axis denotes the accuracy. Each color denotes the $nn$.}
    \label{UMAP}
\end{figure}

\begin{figure}[tb]
    \centering
    \includegraphics[width=40mm]{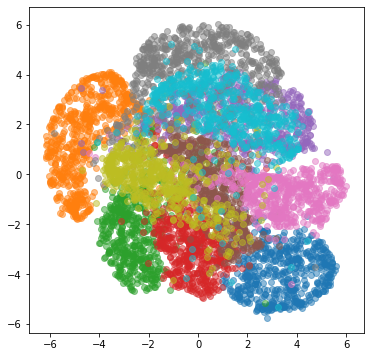}
    \includegraphics[width=42.5mm]{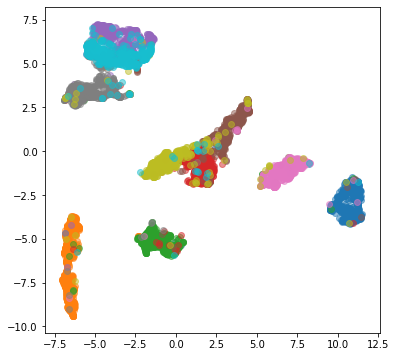}
    (a) SNE [G.Hinton] \hspace*{11mm} (b) UMAP [L.Mclnnes]\\
    \includegraphics[width=40mm]{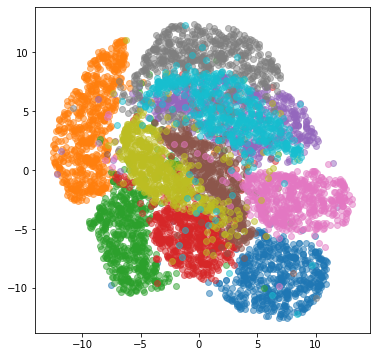}
    \includegraphics[width=40mm]{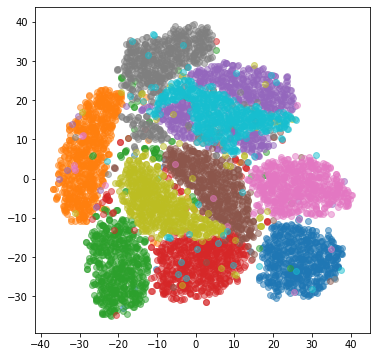} \\
    (c) $q=1.1$ \hspace*{25mm} (d) $q=1.5$\\
    \includegraphics[width=40mm]{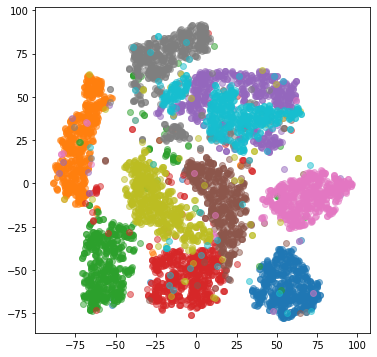}
    \includegraphics[width=40mm]{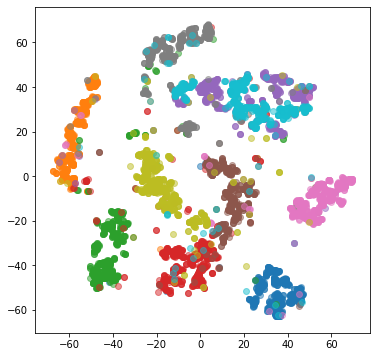}\\
    (e) $q=2.0$ (t-SNE) \hspace*{16mm} (f) $q=2.5$\\
    \includegraphics[width=\linewidth]{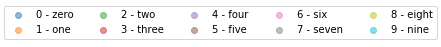}
    \caption{Visualization of MNIST dataset in the 2-D embedded spaces.}
    \label{mnist}
\end{figure}


\begin{figure}[tb]
    \centering
    \includegraphics[width=40mm]{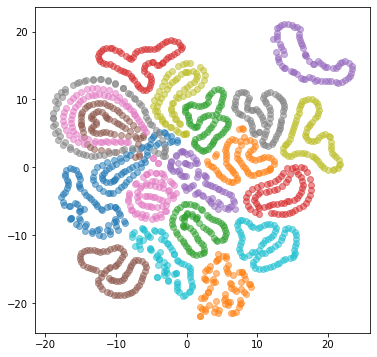}
    \includegraphics[width=40mm]{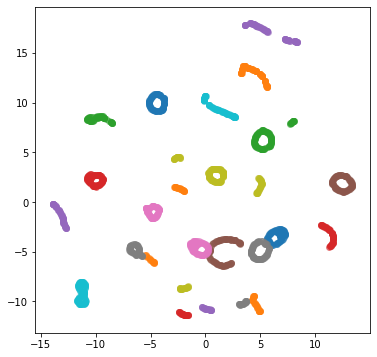} \\
    (a) SNE [G.Hinton] \hspace*{11mm} (b) UMAP [L.Mclnnes]\\
    \includegraphics[width=40mm]{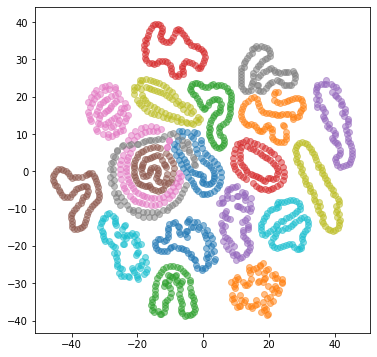}
    \includegraphics[width=40mm]{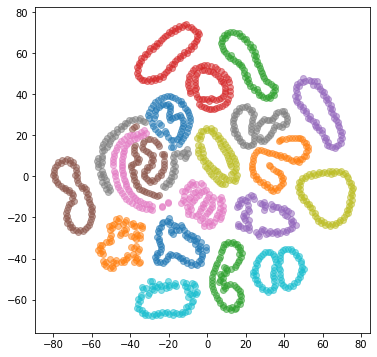} \\
    (c) $q=1.1$ \hspace*{25mm} (d) $q=1.5$\\
    \includegraphics[width=40mm]{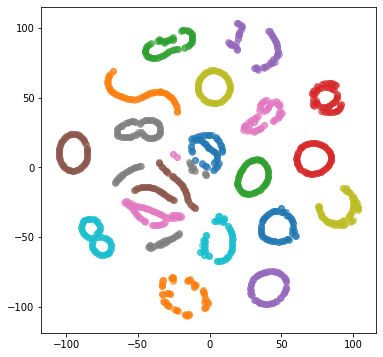}
    \includegraphics[width=40mm]{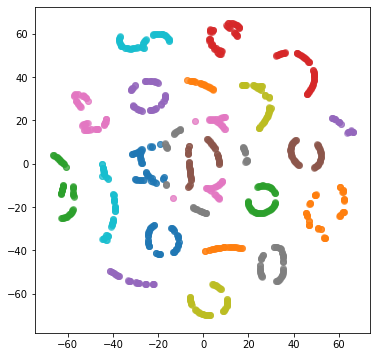}\\
    (e) $q=2.0$ (t-SNE) \hspace*{16mm} (f) $q=2.5$\\
    \includegraphics[width=\linewidth]{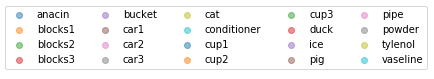}
    \caption{Visualization of COIL-20 dataset in the 2-D embedded spaces.}
    \label{coil}
\end{figure}

\begin{figure}[t]
    \centering
    \includegraphics[width=40mm]{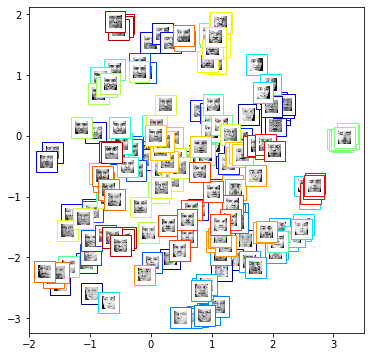}
    \includegraphics[width=40mm]{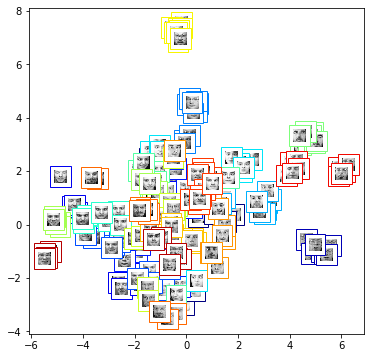} \\
    (a) SNE [G.Hinton] \hspace*{11mm} (b) UMAP [L.Mclnnes]\\
    \includegraphics[width=40mm]{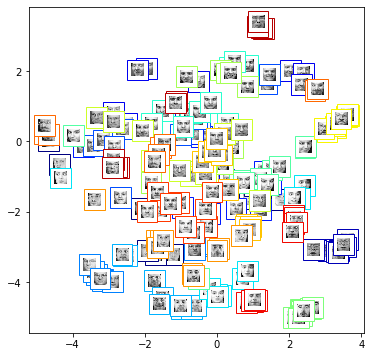}
    \includegraphics[width=40mm]{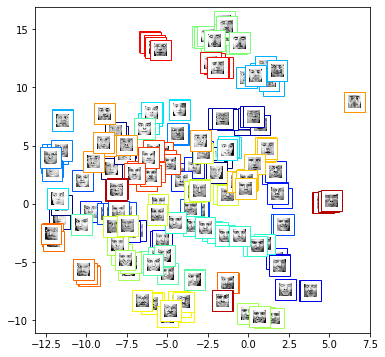}\\
    (c) $q=1.1$ \hspace*{25mm} (d) $q=1.5$\\
    \includegraphics[width=40mm]{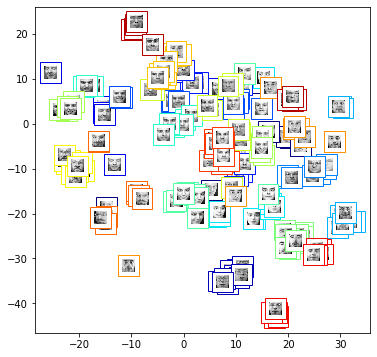}
    \includegraphics[width=40mm]{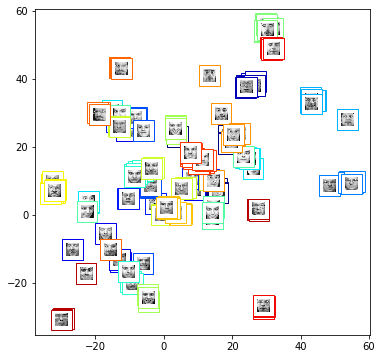}\\
    (e) $q=2.0$ \hspace*{25mm} (f) $q=2.5$\\
    \caption{Visualization of OlivettiFaces in the 2-D embedded spaces.}
    \label{olivetti}
\end{figure}

\begin{figure}[tb]
    \centering
    \includegraphics[width=43mm]{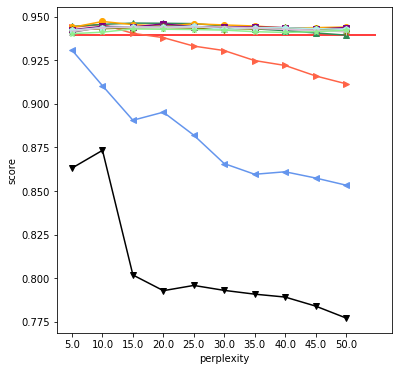}
    \includegraphics[width=43mm]{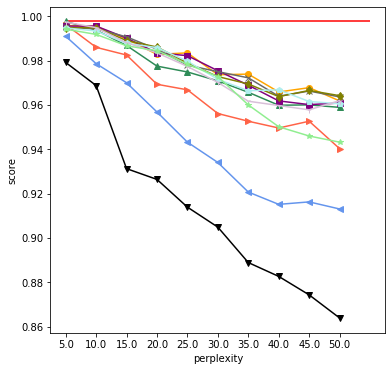}\\
    (a) MNIST \hspace*{25mm}
    (b) COIL-20\\
    \includegraphics[width=43mm]{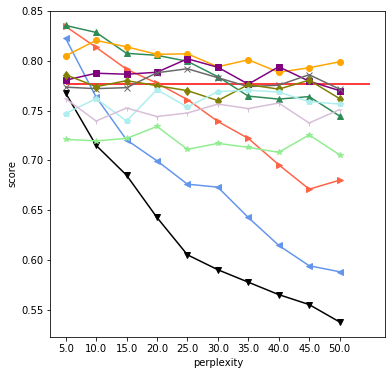}
    \includegraphics[width=43mm]{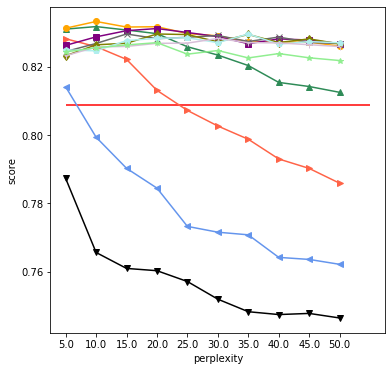}\\
    (c) OlivettiFaces \hspace*{25mm}
    (d) FashionMNIST\\
    \includegraphics[width=\linewidth]{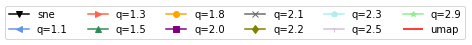}
    \caption{These graphs show the relation between the classification accuracy by the k-NN classifier($k=10$) in the embedded space and the perplexity. The x-axis denotes the perplexity. The y-axis denotes the classification accuracy. The results of different parameters $q$ are shown in different colors.}
    \label{graphresults}
\end{figure}

\begin{table}[tb]
    \begin{center}
        \caption{The $Q_{local}$ of the scale-independent quality criteria.}
          \begin{tabular}{p{1.2cm}||c|c|c|c} \hline
                & MNIST & COIL-20 & OlivettiFaces & FashionMNIST \\ \hline \hline
            SNE [G.Hinton] & 0.3861 & 0.8328 & 0.6888 & 0.5672 \\ \hline
            $q=1.1$ & 0.4557 & 0.8561 & 0.8064 & 0.5794 \\ \hline
            $q=1.3$ & 0.4788 & 0.8736 & 0.8640 & 0.5809 \\ \hline
            $q=1.5$ & 0.6518 & \textbf{0.8848} & \textbf{0.8770} & 0.6204 \\ \hline
            $q=1.8$ & \textbf{0.6990} & 0.8788 & 0.8750 & \textbf{0.6663} \\ \hline
            $q=2.0$ (t-SNE) & 0.6735 & 0.8534 & 0.8585 & 0.6359 \\ \hline
            $q=2.1$ & 0.6450 & 0.8376 & 0.8440 & 0.6069 \\ \hline
            $q=2.2$ & 0.6178 & 0.8153 & 0.8291 & 0.5805 \\ \hline
            $q=2.3$ & 0.5880 & 0.7955 & 0.8269 & 0.5576 \\ \hline
            $q=2.5$ & 0.5431 & 0.7673 & 0.7995 & 0.5529 \\ \hline
            $q=2.9$ & 0.4814 & 0.7162 & 0.6242 & 0.5412 \\ \hline
            UMAP [L.Mclnnes] & 0.4686 & 0.6772 & 0.5598 & 0.5557 \\ \hline
          \end{tabular}
        \label{qtable}
    \end{center}
\end{table}

\begin{figure}[tb]
    \centering
    \includegraphics[width=40mm]{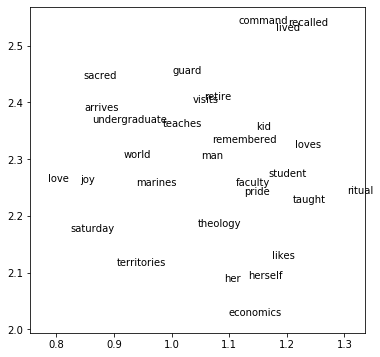}
    \includegraphics[width=40mm]{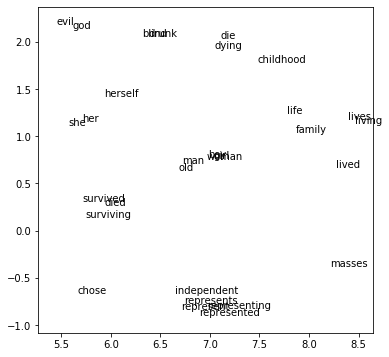}\\
    (a) $q=1.1$ \hspace*{25mm}
    (b) $q=1.5$\\
    \includegraphics[width=40mm]{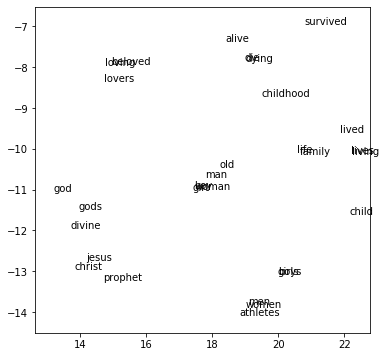}
    \includegraphics[width=40mm]{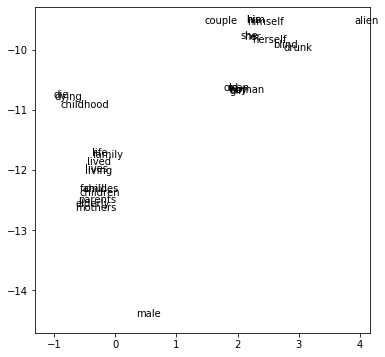}\\
    (c) $q=2.0$ \hspace*{25mm}
    (d) $q=2.5$\\
    \caption{Visualization of Glove data in the 2-dimensional embedded space.}
    \label{glove}
\end{figure}

\begin{table}[tb]
    \begin{center}
        \caption{Top 25 words around "man" in order from closer words by using q-SNE with each parameter $q$.}
          \begin{tabular}{c|c|c|c} \hline
             $q=1.1$ & $q=1.5$ & $q=2.0$ & $q=2.5$ \\ \hline \hline
             remembered     & old           & boy           & woman \\
             faculty        & woman         & woman         & boy \\
             teaches        & boy           & old           & girl \\
             pride          & girl          & girl          & old \\
             visits         & died          & childhood     & her \\
             retire         & herself       & life          & she \\
             kid            & survived      & family        & herself \\
             theology       & her           & dying         & blind \\
             student        & surviving     & die           & drunk \\
             marines        & life          & boys          & himself \\
             world          & family        & girls         & him \\
             guard          & she           & men           & his \\
             loves          & dying         & alive         & couple \\
             taught         & childhood     & women         & alien \\
             undergraduate  & independent   & athletes      & life \\
             joy            & die           & lovers        & family \\
             likes          & drunk         & gods          & lived \\
             arrives        & blind         & beloved       & lives \\
             her            & represents    & prophet       & living \\
             herself        & represent     & loving        & childhood \\
             territories    & lived         & jesus         & dying \\
             command        & representing  & lived         & die \\
             sacred         & represented   & divine        & child \\
             saturday       & lives         & lives         & families \\
             lived          & chose         & child         & children \\
             \hline
          \end{tabular}
          \label{wordtable}
    \end{center}
\end{table}

\subsection{Visualization of Image Data}

Next, we will show the results of visualization of image datasets MNIST, COIL-20, and OlivettiFaces.
To compare the visualization, we use the SNE and UMAP.
The MNIST dataset contains 60,000 grayscale images of handwritten digits.
The size of each image is $28\times 28$ pixels.
For the visualization, 6,000 images were randomly selected from the dataset.
The COIL-20 dataset contains 1,440 grayscale images of 20 objects.
The objects were placed on a motorized turntable against a black background.
The turntable was rotated through 360 degrees to vary object pose with respect to a fixed camera. 
Images of the objects were taken at pose intervals of 5 degrees.
The size of each image is $128\times 128$ pixels.
The OlivettiFaces dataset contains 400 grayscale face images taken from 40 persons.
The size of each image is $92\times 112$ pixels.

To reduce the dimension of the original high-dimensional space, we applied the Principle Component Analysis (PCA)\cite{pearson1901liii} and obtained the vectors of 30-dimension for each sample in the datasets.
These vectors were used as the input of the SNE, UMAP, and q-SNE.

Then the 2-dimensional embedded space is constructed for each dataset by using the SNE, UMAP and proposed q-SNE with different parameters $q=1.1$, $q=1.5$, $q=2.0$(t-SNE), and $q=2.5$.
To calculate the joint probability $p_{ij}$ in the original high-dimensional space, 
the perplexity is set to 30 for MNIST, 5 for COIL-20, and 25 for OlivettiFaces, respectively.

To find the best visualization of UMAP, we investigated hyperparameters by using the classification accuracy of the k-Nearest Neighbors (k-NN) classifier on the embedding space.
The $k$ of the k-NN classifier was set to $10$.
We fix $d$ and $n-epochs$ to 2 and 200 respectively.
We set $nn$ and $min\_dist$ to 5-50 and 0.001-0.5 respectively.
And we compute the classification accuracy averaged 5 trials with different seeds.
In Fig.\ref{UMAP}, we show the classification accuracy on the embedding space.
For visualization, we used the hyperparameters that are the best score.

Fig.\ref{mnist}, Fig.\ref{coil}, and Fig.\ref{olivetti} show the plot of the 2-dimensional embedded spaces for MNIST, COIL-20, and OlivettiFaces.
And we show the images of 20 persons in Fig.\ref{olivetti}.
It is noticed that the clusters shrink as the value of the parameter $q$ increases.
Since the proposed q-SNE is the same as the t-SNE when $q=2.0$ and is also close to the SNE when $q=1.1$, it is possible to control the degree of the compactness of clusters of the samples in the visualization by changing the parameter $q$ including the visualizations by the SNE or the t-SNE.
It is expected that this flexibility can make use as a visualization tool.

\subsection{Classification by k-NN classifier in the 2-D Embedded Space}

To evaluate the embedded 2-dimensional spaces constructed by the SNE, UMAP , and proposed q-SNE, we have performed the experiments on classification by the k-Nearest Neighbors (k-NN) classifier on the embedded space.
The $k$ of the k-NN classifier was set to $10$ and the classification accuracy between the predicted labels and the teacher labels is calculated as the average of 5 trials with different seeds.
Since the variance $\sigma$ to calculate the joint probability $p_{ij}$ in the high-dimensional space affects to the embedding, the classification accuracy is measured by changing the perplexity from $5$ to $50$.
The average classification accuracy is calculated for different parameters $q$.
The score of UMAP is chosen as the vest score from Fig. \ref{UMAP}.

The results for MNIST, COIL-20, OlivettiFaces, and  FashionMNIST datasets are shown in Fig. \ref{graphresults}.
The FashionMNIST contains 60,000 grayscale images of clothes.
The size of each image is $28\times 28$ pixels, which is the same as MNIST, and 6,000 images are randomly selected from the dataset and used for classification experiments.

From Fig.\ref{graphresults}, it is noticed that the average classification accuracy for the case of $q=1.8$ is almost the best for all datasets.
Our proposal q-SNE is better than the UMAP for classification.
Furthermore, we can consider that this classification accuracy result indicates the optimal value of q for each dataset.

Morover, we evaluated the embedding space by using the scale-independent quality criteria \cite{lee2010scale}.
We show the $Q_{local}$ averaged 5 trials with different seeds corresponding to each data set and each q value in TABLE \ref{qtable}.
The higher the value of $Q_{local}$, the better the embedding.
According to this TABLE \ref{qtable}, The case of $q=1.8$ or $q=1.5$ of the q-SNE is the best for all datasets.


\subsection{Embedding of Word Data}

We did the experiments on the Glove dataset as word dataset.
In our experiments, we visualize high-dimensional data points on 2-dimensional embedded space by using q-SNE with few different parameter $q$.

The GloVe dataset contains 400,000 words which has a $300$ dimensions vector.
For our experiments, we randomly choose 10,000 words in the dataset.

We show the 30 words close to the word "man" in the 2-dimensional embedded space for the cases with parameters $q=1.1$, $q=1.5$, $q=2.0$, and $q=2.5$ in Fig.\ref{glove}.
And we show the top 25 words close to the word "man" in TABLE \ref{wordtable}.

\section{Conclusion}
In this paper, we propose the novel technique called q-SNE for dimensionality reduction.
The q-SNE uses the q-Gaussian distribution which can recover the Gaussian distribution and the t-distribution by setting the parameter $q$ in the q-Gaussian distribution.

Through our experiments, the proposed qSNE can produce results similar to SNE or the same as tSNE by changing the parameter $q$.
And the q-SNE is better than the SNE, t-SNE and, UMAP by the accuracy of the k-NN classifier on MNIST, COIL-20, OlivettiFaces, and  FashionMNIST.
We can know that the q-SNE makes it possible to find the best visualization or best classification by choosing the parameter $q$.


\bibliography{main}
\bibliographystyle{unsrt}

\end{document}